\documentclass[10 pt,a4paper]{article} 
\usepackage[margin=1.5in]{geometry}
\usepackage{graphicx} 
\usepackage{color}
\usepackage{amsmath}
\usepackage{amssymb} 
\usepackage{underscore}
\usepackage{fancyvrb}
\usepackage{algorithmic}

\title{Model-based Test Generation for Robotic Software: Automata versus Belief-Desire-Intention Agents}

\author{Dejanira Araiza-Illan\footnote{Department of Computer Science and Bristol Robotics Laboratory, University of Bristol, UK. E-mail: {\tt\footnotesize dejanira.araizaillan@bristol.ac.uk}}, Anthony G. Pipe\footnote{Faculty of Engineering Technology and the Bristol Robotics Laboratory, University of the West of England, Bristol, UK. E-mail:~{\tt\footnotesize tony.pipe@brl.ac.uk}} and Kerstin Eder\footnote{Department of Computer Science and Bristol Robotics Laboratory, University of Bristol, UK. E-mail: {\tt\footnotesize kerstin.eder@bristol.ac.uk}} \footnote{This work was supported by the EPSRC grants EP/K006320/1 and EP/K006223/1, part of the project ``Trustworthy Robotic Assistants''.}}

\date{}

\begin{document}

\maketitle

\begin{abstract}
Robotic code needs to be verified to ensure its safety and functional correctness, especially when the robot is interacting with people.
Testing real code in simulation is a viable option. 
However, generating tests that cover rare scenarios, as well as exercising most of the code, is a challenge amplified by the complexity of the interactions between the environment and the software. 
Model-based test generation methods can automate otherwise manual processes and facilitate reaching rare scenarios during testing. 
In this paper, we compare using Belief-Desire-Intention (BDI) agents as models for test generation with more conventional automata-based techniques that exploit model checking, in terms of practicality, performance, transferability to different scenarios, and exploration (`coverage'), through two case studies: a cooperative manufacturing task, and a home care scenario. 
The results highlight the advantages of using BDI agents for test generation.
BDI agents naturally emulate the agency present in Human-Robot Interactions (HRIs), and are thus more expressive than automata. 
The performance of the BDI-based test generation is at least as high, and the achieved coverage is higher or equivalent, compared to test generation based on model checking automata.
\end{abstract}

\section{INTRODUCTION}

As robot software designers, we must demonstrate the safety and functional soundness of robots that interact closely with people, if these technologies are to become viable commercial products~\cite{ROMAN14}. 
Beyond the elimination of runtime errors, a robot's code must be verified and validated at a functional level, with respect to hardware and other software components, and interactions with the environment, including with people. 
The interaction of all these elements introduces complexity and concurrency, and thus the possibility of unexpected and undesirable behaviours~\cite{Micskei2012,CDV2015}.

Robot high-level behaviours and control code have been verified via model checking, either by hand-crafting an abstract model of the code or behaviours (as in~\cite{webster14formalshort}), or by automated translations from code, often restricted to a limited subset of the language, into models or model checking languages (as in~\cite{BordiniVMAPMC}). 
These models might require subsequent abstraction processes~\cite{Clarke2001} that remove detail from the original code, to make verification feasible~\cite{Patelli2016}. 
Furthermore, the equivalence between the original code and the model needs to be demonstrated, for the verification results to be considered truthful (as it is done in counter-example guided abstraction refinement~\cite{Clarke2003}). 
Alternatively, robots' code can be tested directly, at the cost of verification not being exhaustive. 
An advantage of testing is that realistic components, such as emulated or real hardware (hardware-in-the-loop) and users (human-in-the-loop) can be added to the testing environment ~\cite{Petters2008,Mossige2014,Pinho2014}.

Formal methods explore models fully automatically and exhaustively. They have been used for model-based test generation~\cite{Fraser2009}, reducing the need for writing tests manually.
In model-based testing, a model of the system under test or the testing goals is derived first, followed by its traversal to generate tests based on witness traces or paths~\cite{Utting2012}. 
The new challenge in testing robotic code is finding suitable models for test generation, i.e.\ models that capture the interactions between the robot, human and environment in an effective and natural way.

This paper compares Belief-Desire-Intention (BDI) agents with a conventional form of model-based test generation, in terms of practicality, performance, transferability to different scenarios, and exploration, within the context of software for robots in Human-Robot Interactions (HRIs), investigating the following research questions:
\begin{enumerate}
\item How does the use of BDI agents compare with model checking timed automata (TA), for model-based test generation in the HRI domain?
\item Are BDI models useful to generate effective and efficient tests for different types of HRI applications?
\end{enumerate}

We use two case studies to evaluate the practicality, performance, transferability and exploration capabilities of BDI-based test generation vs.\ test generation based on model checking TA: a human-robot cooperative manufacturing task, and a home care assistant scenario.
The generated tests were run in a simulation of the scenarios, to gather statistics on coverage of code (executed lines), safety and functional requirements (monitored during execution), and combinations of human-robot actions (denominated cross-product, Cartesian product, or situational coverage~\cite{Alexander2015,TAROS2016}). 
Our results demonstrate that BDI agents are effective and transferable models for test generation in the HRI domain.
Compared to traditional test generation by model checking automata, BDI agents achieve better or similar code, requirement and cross-product coverage and are stronger at finding requirement violations.
Also, BDI agents are easy to implement, and can be explored quickly.

\section{RELATED WORK} \label{sc:relatedwork}

In our previous work, we presented a simulation-based method to test real, high-level robot control code in an effective and scalable manner~\cite{CDV2015,TAROS2016}.
Automation of the testing process and a systematic exploration of the code under test within HRI scenarios was achieved through Coverage-Driven Verification (CDV), a method that guides the generation of tests, according to feedback from coverage metrics~\cite{Pizialli2004}. 
In~\cite{CDV2015,TAROS2016}, we illustrated how a CDV testbench, comprising a test generator, driver, self-checker and coverage collector, can be integrated into a simulator running in the Robot Operating System (ROS)\footnote{http://www.ros.org/} framework and Gazebo\footnote{http://gazebosim.org/}.
In this paper we focus on effective and efficient test generation.

In many robotics applications, test generation has been needed only for stimulating dedicated and simple components (equivalent to unit testing), such as choosing from a set of inputs for a controller~\cite{Kim2006}, or generating a timing sequence for activating individual controllers~\cite{Mossige2014}. 
For these applications, random data generation or sampling~\cite{Nie2011} might suffice to explore the state space or data ranges~\cite{Kim2006}, along with alternatives such as constraint solving or optimization techniques~\cite{Mossige2014}. 
When testing a full robot system, however, the orchestration of different timed sequences in parallel (e.g.\ for emulated hardware components), coupled with several tasks of data instantiation (e.g.\ for sensor readings), is more complex.
Sophisticated model-based approaches, such as those presented in this paper, offer a practical and viable solution for complex test generation problems~\cite{Fraser2009,Utting2012,Diasneto2007}. 
A model-based approach can be used in a hierarchical manner in order to coordinate lower-level random data generation and optimization with more complex, higher-level test generation tasks. A two layered approach is proposed in~\cite{CDV2015,TAROS2016}.

Many languages and formalisms have been proposed for generic software model-based test generation~\cite{Shafique2015}, e.g.\ UML and process algebras for concurrency~\cite{Lill2012}, or Lustre and MATLAB/Simulink for data flow~\cite{Utting2012}. Their suitability for the HRI domain, in terms of capturing realistic and uncertain environments with people, is yet to be determined~\cite{Namin2010}.  
Also, deriving models automatically from generic code (e.g.\ Python and C++ in ROS), or from user requirements, remains a challenge. 
BDI agents~\cite{Agentspeakbook} have been used successfully to model decision making in autonomous robots~\cite{Dennis2016}. 
Because BDI agents naturally reflect agency, they are also ideal to model the agency present in the robot's environment (e.g.\ people).
In~\cite{MORSE2016}, we have shown how to use BDI agents for model-based test generation.

If a model is available (e.g.\ a functional modular description in~\cite{Abdellatif2012}), as in model-based software engineering, the verification of the software with respect to functional requirements captured in the model can be performed at design time. 
Code can then be generated (e.g.\ refined) from the verified model. 
However, mechanisms such as certified code generators are needed to ensure the code is equivalent to the model and thus meets its requirements~\cite{Toom2010}.

\section{CASE STUDIES}\label{sc:casestudy}

Our two case studies are a cooperative manufacturing scenario and a basic home care assistant scenario.

\subsection{Cooperative Manufacturing Task}

We used the scenario we presented in~\cite{MORSE2016}, where a human and a robot collaborate to jointly assemble a table.
The robot, BERT2~\cite{lenz2010bert2}, should, when asked, hand over legs to the person, one at a time. 
A table is completed when four legs have been handed over successfully within a time threshold. 
For this paper, the code and simulator in~\cite{MORSE2016} were slightly modified.

A handover starts with a voice command from the person to the robot, requesting a table leg.  
The robot then picks up a leg, holds it out to the human, and signals for the human to take it. 
The human issues another voice command, indicating readiness to take the leg. Then, the robot makes a decision to release the leg or not, within a time threshold, based on a combination of three sensors: ``pressure'' (the human is holding the leg); ``location'' (the person's hand is near the leg); and ``gaze'' (the person is looking at the leg). 

All sensor combinations are captured by the Cartesian product of ``gaze'', ``pressure'' and ``location'' readings, $(g,p,l)\in G \times P \times L$. Each sensor reading is classified into $G=P=L=\{\bar{1},1\}$, with $1$ indicating the human `is ready', and $\bar{1}$ for any other value. 
If the human is deemed ready, indicated by $GPL=(1,1,1)$, the robot should decide to release the leg. Otherwise, the robot should not release the leg. 
The robot will time out while waiting for either a voice command, or the sensor readings. 
The human can `get bored' and disengages from the collaboration, aborting the handover. 

A ROS `node' (code under test) with 264 lines of executable code in Python implements the robot's control (e.g.\ calls the kinematic planner MoveIt! and reads the sensor inputs). The code was structured as a finite-state machine (FSM) using SMACH~\cite{SMACH}. This allows an efficient implementation of control flow. 
The FSM has 14 states and 22 transitions.

\subsubsection*{Requirements}\label{ssc:requirements}

We considered the following selected set of safety and functional requirements from~\cite{MORSE2016} and the standards ISO~13482 (personal care robots), ISO~15066 (collaborative robots) and ISO~10218 (industrial robots):

\begin{enumerate}
\item The robot shall always discard or release a leg within a time threshold, whenever it reaches the sensing stage and determines the human is ready or not. (functional)
\item If the gaze, pressure and location indicate the human is ready, then a leg shall be released. (functional)
\item If the gaze, pressure or location indicate the human is not ready, then a leg shall not be released. (functional) 
\item The robot shall always discard or release a leg, when activated. (functional) 
\item The robot shall not close its hand when the human is too close. (safety)
\item The robot shall start and work in restricted joint speed of less than 0.25 rad/s. (safety)
\end{enumerate}

\subsection{Home Care Assistant}

A TIAGo robot\footnote{http://tiago.pal-robotics.com/} operates in a flat.
%the lower floor of a house.
%
The robot is in charge of taking care of a person with limited mobility, by bringing food to the table (`feed'), clearing the table (`clean'), checking the fridge door (`fridge'), and checking the sink taps (`sink'). 
The robot's code invokes motion sequences, assembled from primitives such as `go to fridge', `go to table', or `open the gripper', to obey commands.
Whenever the person asks the robot to execute a command that is not in the list of known ones, the robot will not move. 
The robot moves to a default location, denominated `recharge', after completing a command, and should remain there until the person asks it to do something else.
We assume the person will not ask the robot to perform more than three feasible tasks within a 10 minute interval. 

\begin{figure}[t]
\centering
\includegraphics[scale=0.3]{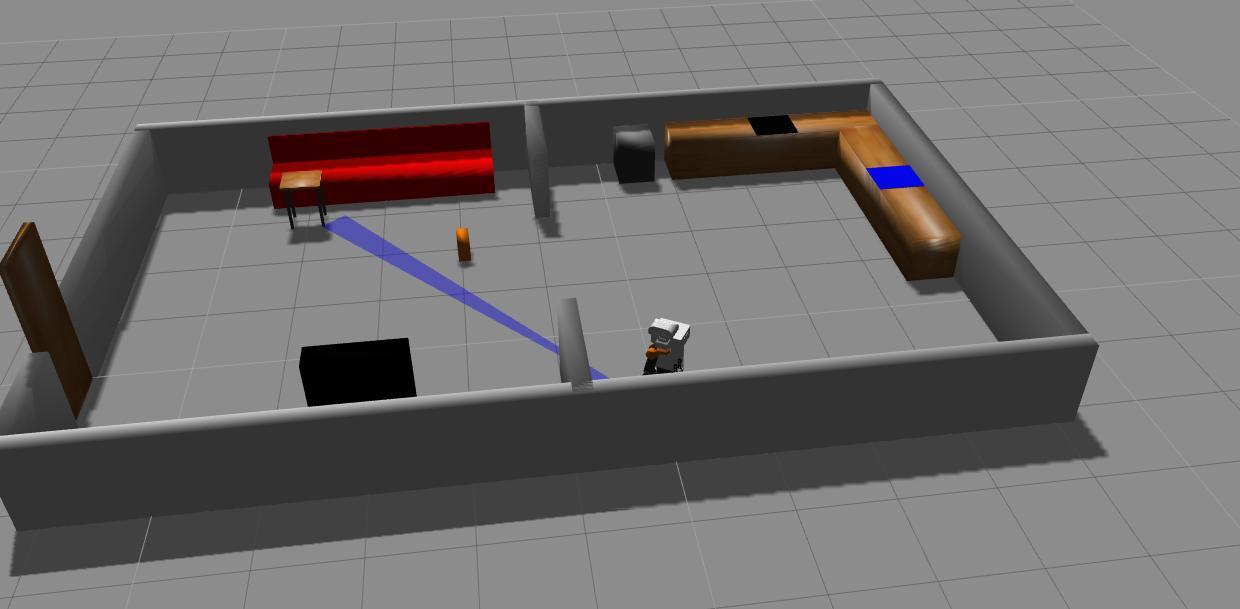}
\vspace*{-3mm}\caption{TIAGo in the simulated living space with a dog (orange).}
\label{tiago}
\end{figure}

A small dog cohabits the operational space. To avoid dangerous collisions with the dog, the robot checks the readings from the laser scan and stops if any object is too close (within a proximity of 20 cm). Figure~\ref{tiago} shows the simulated environment and the robot.

A ROS `node' (code under test) with 265 executable lines of code in Python implements the robot's control (e.g.\ motion towards a goal, listening for human commands, and collision avoidance). We used the ROS infrastructure provided in TIAGo's repository\footnote{http://wiki.ros.org/Robots/TIAGo}. 
The code contains 5 FSMs within the code (to execute each one of the location to location motions, e.g.\ table to sink, or sink to recharge station) with 5, 6, 3, 3 and 2 states, and 4, 5, 2, 2, and 1 transitions, respectively.

\subsubsection*{Requirements}

We considered the following requirements, also inspired by safety standards:

\begin{enumerate}
\item If the robot gets food from the fridge, it shall eventually bring it to the table. (functional)
\item If the robot is idle waiting for the next order, it shall go to or remain in the recharge station. (functional)
\item The robot shall start and operate in restricted motion speed of less than 0.25 m/s. (safety)
\item The robot shall not stay put (unless in the recharge station) for a long period of time. (functional) 
\end{enumerate}

\subsection{Simulator in ROS and Gazebo}\label{ssc:simulator}

A 3-D model was built for each scenario in Gazebo. A simulation in Gazebo is controlled by the robot's code on one hand, and by code modelling the human, sensors, objects (legs and the dog), and others (MoveIt!) on the other hand (the environment), all of them running in ROS. 
The simulators for both scenarios are available online\footnote{https://github.com/robosafe/mc-vs-bdi}.

CDV testbench components (a driver, a checker, and coverage collection) were extended from the ROS infrastructure previously developed in~\cite{CDV2015,TAROS2016}, for each case study. 
We implemented assertion monitors for each one of the requirements in the two case studies as self-checkers, reusing monitors where possible, e.g. for Req.~1 in both scenarios. 
The monitors are executed in parallel with the code in the simulation, and restarted every time a test is run. 
Coverage collection is logged automatically if a monitor is triggered. A post-processing step collects statistics on which monitors were triggered by a test.

We implemented code coverage via the `coverage' Python module\footnote{http://coverage.readthedocs.org/en/coverage-4.1b2/}, which  automatically records the executed lines of code in log files. 
This provides branch coverage~\cite{Pizialli2004}.

We also implemented cross- or Cartesian product coverage, capturing interactions of the robot and its environment, ($Human \times Robot$), deemed to be fundamental for the scenarios. For example, combinations of requesting 1 to 4 legs, and the robot deciding to release all, some or none in the manufacturing task. 
This coverage is collected offline after the test runs by traversing simulation log files.
In the manufacturing scenario, we collapsed the cross-product of all interactions\footnote{There are 6500 coverage tasks in the cross-product.} into 14 subgroups as shown in Table~\ref{fig:cross-product}. This captures that the human could ask for 1 to 4 legs, none or get bored, and the robot's sensor readings have 8 possible values, or the robot can time out, for each requested leg. 
In the home care scenario, we also collapsed the cross-product into 6 subgroups\footnote{From more than 250 coverage tasks in the cross-product, with a maximum of 3 commands given per test, which the robot might or might not complete.}, as shown in Table~\ref{fig:cross-product}. We capture 5 possible requests from the human (including one issuing any invalid request), and outcomes for the robot where none, at least one or at least two of the specified requests were completed successfully.

The test generator is run before the simulation (offline test generation). 
We used our previously proposed two-tiered test generation process~\cite{CDV2015}, where abstract tests in the form of timed sequences are generated first, and then valid data is instantiated. 
A test stimulates the environment the robot interacts with in the simulation. This engages the robot in interactions, thereby stimulating the code under test. 
Example tests can be found in~\cite{TAROS2016,MORSE2016}.
We instantiated and extended our previous implementations of test generation by pseudorandom sampling, model checking TA~\cite{CDV2015,TAROS2016} and BDI agents~\cite{MORSE2016}, for the case studies in this paper. %, which we describe next. 

\section{MODEL-BASED TEST GENERATION}\label{sc:modelbased}

We describe two types of model-based test generation, using BDI agents~\cite{MORSE2016} and model checking TA~\cite{CDV2015,TAROS2016}, along with a baseline: pseudorandom test generation. 
In model-based approaches, a model of the system or its requirements is assembled first and then explored to produce tests. 
We use model-based approaches to produce abstract tests that will indirectly stimulate the robot's code in simulation by stimulating the environment that the code interacts with, instead of stimulating the code directly.

\subsection{BDI Agent Models and Exploration}\label{sc:bdi}

BDI is an intelligent or rational agent architecture for multi-agent systems. BDI agents model human practical reasoning,
in terms of `beliefs' (knowledge about the world and the agents), `desires' (goals to fulfil), and `intentions' (plans to execute in order to achieve the goals, according to the available knowledge)~\cite{Agentspeakbook}. 
Recently, we have shown that BDI agents are well suited to model rational, human-like decision making in HRI, for test generation purposes~\cite{MORSE2016}. 

We employed the Jason interpreter, where agents are expressed in the AgentSpeak language. 
An agent is defined by initial beliefs (the initial knowledge) and desires (initial goals), and a library of `plans' (actions to execute to fulfil the desires, according to the beliefs). 
A plan has a `head', formed by an expression about beliefs (a `context') serving as a guard for plan selection, and a `body' or set of actions to execute. 
New beliefs appear during the agents' execution, can be sent by other agents, or are a result of the execution of plans (self-beliefs)~\cite{Agentspeakbook}.

We model an HRI scenario using a set of BDI agents, representing the robot's code, sensors, actuators, and its environment.
Then, we add BDI verification agents that control the execution of the HRI agents, by triggering (sending) beliefs to activate plans and create new desires in the other agents, which in turn may lead to the triggering of new plans, and so on. 
A set of beliefs for the verification agents to send to other agents is chosen, and then the multi-agent system is executed once. 
Each system execution with a different set of beliefs will activate a corresponding sequence of plans in the agents.
This execution (a set of chosen and executed plans) is recorded and used to generate an abstract test (a sequence of `actions' according to the recorded plans).
We extract the environment components from an abstract test to stimulate the robot's code indirectly.

\subsubsection{Model for the Cooperative Manufacturing Task}
We reused the BDI model in~\cite{MORSE2016}, with minor modifications.
The model consists of four agents: the robot's code, the human, the sensors (as a single agent), and the verification one. 
The verification agent makes the human agent send activation signals to both the robot's code (voice commands) and the sensors agent. 
There are a total of 38 possible beliefs for the verification agent, including, e.g., requesting 1 to 4 legs, readings for the three sensors for each leg, and the human getting bored. 
The sensors agent transmits readings of either $1$ or $\bar{1}$ to the robot's code agent. 
The robot's code agent has a similar structure to the FSM in the real code, interacting only with the human and sensors agents through beliefs. 

\subsubsection{Model for the Home Care Assistant}
Our model consists of five agents: the robot's code, the human, the dog, the sensor (for collision avoidance), and the verification one. 
The verification agent selects the requests that the human agent communicates to the robot's code agent, one at a time.
The dog agent can opt to collide with TIAGo or not. This is then perceived by the sensor agent, which transfers this information to the robot's code agent.  
The robot's code agent is based on an FSM that is similar to the one used in the real code.
There are 5 possible beliefs for the verification agent to control the human, comprising the four available requests and an extra one representing any other invalid request. 

\subsection{Timed Automata Models and Model Checking}\label{sc:pta}

Model checking is the exhaustive exploration of a model to determine whether a logic property is satisfied or not. 
Traces of examples or counterexamples are provided as evidence of satisfaction or proof of violation, respectively. 
Model checking applied to models of the code or high-level system functionality can be exploited for model-based test generation, where these traces are used to derive tests~\cite{Fraser2009,Utting2012}.

In~\cite{CDV2015,TAROS2016}, we modelled HRI in terms of TA for the model checker UPPAAL\footnote{http://www.uppaal.org/}. 
Non-determinism allows capturing uncertainty in the human actions, and sensor errors, through the selection of different transitions in the automata.
As robots interacting with humans are expected to fulfil goals in a finite amount of time, the timing counters in the TA allow emulating these timing thresholds. 
The execution of these automata is synchronized by communication events, and `guards' or conditions, to transition from one state to another, according to system variables and events.

To derive tests from the TA, logic properties are formalized manually in TCTL (Timed Computation Tree Logic), and automatically checked by the UPPAAL model checker. 
For example, we would specify that `the robot reaches a specified location within a minute', for the home care scenario. 
Formulating suitable properties to achieve high model coverage requires a good understanding of formal logic, the HRI scenario and the TA models. 
The UPPAAL model checker produces an example if the property is satisfied, comprising sequences of states from all the TA combined. 
To indirectly stimulate the robot's code, we exclude the robot's code contribution from these sequences; what remains is the stimulus used to test the robot's code.

Completeness and correctness of the models was established empirically through step-by-step execution and simulation at development time in Jason and UPPAAL, respectively for BDI agents and TA.
This effort is accounted for in the reported model development time in Section~\ref{sss:perf}.

\subsubsection{Model for the Cooperative Manufacturing Task}

Our model consists of 6 TA, the human, the robot's code, the sensors, and the gaze, pressure, and location selections by the human. 
While the human automaton enacts the activation signals (voice commands), the gaze, pressure and location automata select inputs for the sensors non-deterministically (via variables). 
The sensors automaton reads the variables as $1$ or $\bar{1}$, which are then read by the robot's code automaton to decide whether to release a leg or not. 
The latter has a similar structure to the FSM in the real code. 

\subsubsection{Model for the Home Care Assistant}

Our model consists of 4 TA, the human, the robot's code, the sensor, and the dog.
The sensor automaton determines if the dog is within collision distance or not, according to the choices of the dog automaton. 
The human automaton sets the type of requests for the robot's code automaton, one at a time.
The robot's code automaton, which is similar in structure to the FSMs in the code, executes the requests from the human, whilst considering the sensor readings to avoid collisions.

\subsection{Baseline: Pseudorandom Test Generation}\label{sc:pseudorandom}

As a baseline for comparisons we employed a pseudorandom abstract test generator~\cite{CDV2015,TAROS2016}.
The generator concatenates sequences of `actions' sampled at random from a list of specified ones.
The sequences' length is also chosen pseudorandomly.  
For example, in the home care scenario sequences are assembled from available requests such as `request food' or `request clean'.
In the manufacturing scenario, human actions such as `activate robot' or `choose gaze as OK' are available to be included into sequences.

\section{EXPERIMENTS AND RESULTS}\label{sc:experiments}

\subsection{Experimental setup}
The simulator and testbench were implemented in ROS Indigo and Gazebo 2.2.5. The tests were run on a PC with Intel i5-3230M 2.60\ GHz CPU, 8\ GB of RAM, running Ubuntu 14.04. We used Jason 1.4.2 for the BDI models, and UPPAAL 4.1.19 for model checking. 
All the simulators, code and test generation models used in the experiments, along with the related tests and results data, are available online$^5$.

\subsubsection{Cooperative Manufacturing Assistant}
For model checking TA, as described in Section~\ref{sc:pta}, we manually generated 91 TCTL properties, for which example traces were produced automatically, and abstract tests were extracted. 
These properties covered all the gaze, pressure and location sensor reading combinations, 1 to 4 leg requests, the human getting bored, and the robot timing out while waiting for a signal. 

With the BDI-based method, we generated 131 abstract tests (from a possible total number of $2^{38}$) by specifying constraints for sets of beliefs that covered the same items as the TCTL properties and more, i.e.\ a variety of valid human and robot actions, and an orchestration of the rarest events such as completing 4 legs correctly. 
The generator explores the constrained sets of beliefs automatically, one at a time, over the multi-agent system, following the procedure explained in Section~\ref{sc:bdi}.

Additionally, we generated 160 tests pseudorandomly by sampling from a defined set of human `actions' for the task, as explained in Section~\ref{sc:pseudorandom}.

Each abstract test was concretized at least once by sampling pseudorandomly from valid ranges (i.e.\ parameters were instantiated for variables such as gaze, pressure and location) using as seed the test number, which lead to a total of 160 different concrete tests for each method. 
This process allowed the execution of model-based tests that are equivalent in terms of expected system's abstract functionality, although with different variable instantiations, for both BDI-based and model checking TA methods. 
Each test, once concretized, ran for a maximum of 300 seconds.

\subsubsection{Home Care Assistant}
By model checking TA, we generated 23 TCTL properties and the consequent abstract tests.
These properties covered combinations of 1 to 3 requests for feeding, cleaning, checking the fridge, checking the sink, and any other invalid order.

With the BDI agents, we generated 62 abstract tests by sampling belief sets from a possible total number of $2^5$, to cover the same request combinations as with model checking. 
We discarded 12 tests to get a total of 50, as some of the tests were quite similar (e.g.\ combinations of invalid commands).

Finally, we generated 50 abstract tests pseudorandomly, as explained in Section~\ref{sc:pseudorandom}.

As before, each abstract test was concretized pseudorandomly (at least once in the case of model checking) from valid ranges using as seed the test number, for a total of 50 different concrete tests for each method.
Each test ran for a maximum of 700 seconds.

\subsection{Code Coverage Results}

We expected that the BDI-based method would produce a large number of high-coverage tests quickly, and that both model-based methods would outperform pseudorandom test generation in terms of coverage.

\begin{figure}[!t]
\centering
\includegraphics[scale=0.5]{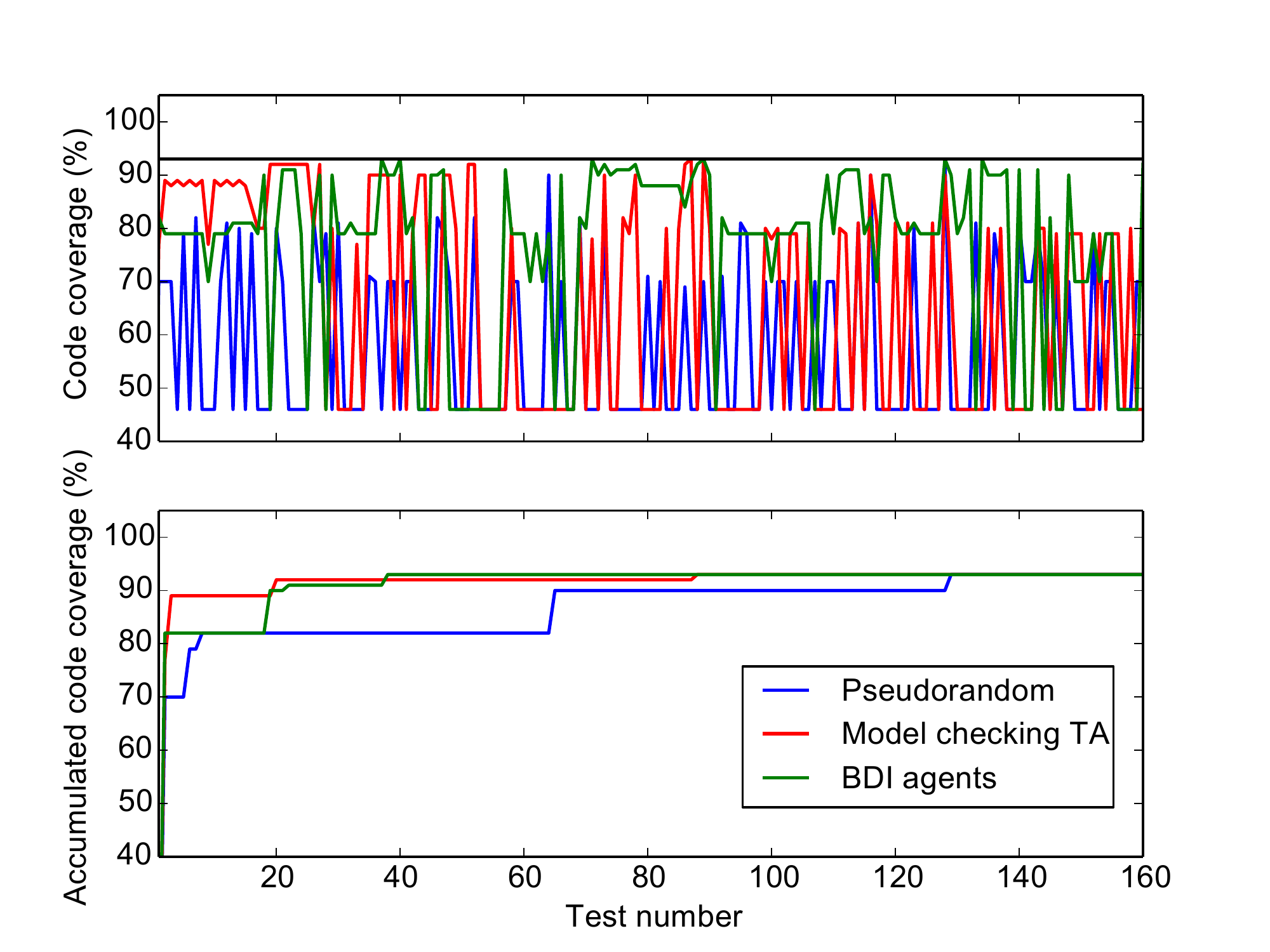}
\vspace*{-5mm}\caption{Code coverage results for the cooperative manufacturing assistant.}
\label{fig:codea}
\end{figure}

\begin{figure}[!t]
\centering
\includegraphics[scale=0.5]{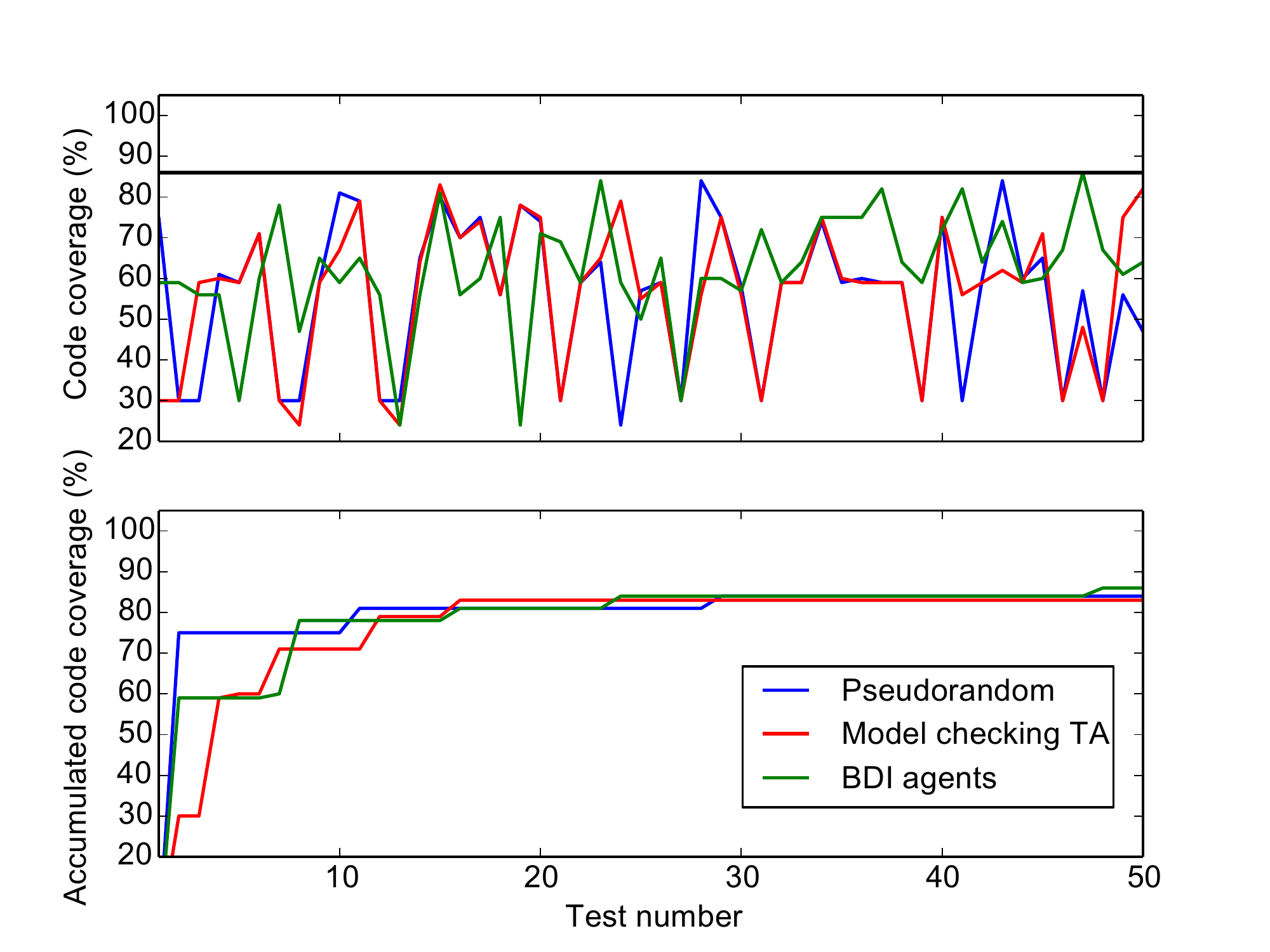}
\vspace*{-5mm}\caption{Code coverage results for the home care assistant.}
\label{fig:codeb}
\end{figure}

Figures~\ref{fig:codea} and~\ref{fig:codeb} show the code coverage percentage reached by each produced test, and the accumulated coverage, for both scenarios. 
In the manufacturing scenario (Fig.~\ref{fig:codea}), tests produced with BDI agents reached high levels of coverage fast (at 92\% of accumulated coverage), and a large number reached the highest coverage possible (92\%), consequently outperforming tests generated pseudorandomly and by model checking TA, in terms of coverage efficiency and effectiveness. 

In the home care scenario (Fig.~\ref{fig:codeb}), tests produced with BDI agents reached the highest coverage results (86\%), outperforming tests generated by model checking TA and pseudorandomly.
Also, Fig.~\ref{fig:codeb} shows that pseudorandomly generated tests start with high accumulated coverage results compared to tests from the other methods, but then this high coverage flattens and tests from the two model-based methods catch up quickly. BDI-based tests outperform the two methods later (at 86\% of accumulated coverage).

\subsection{Assertion Coverage Results}

One of our  motivations for comparing the test generation methods was to establish whether model-based methods would produce tests that achieve higher assertion coverage than pseudorandom test generation. 
Because the models reflect the functional requirements, we expected them to generate tests that trigger the assertion monitors more frequently. 
While this section is focused on assertion coverage, the purpose of assertions is to flag requirement violations. We assess the effectiveness of tests in terms of their ability in finding faults from the triggering of assertion monitors.

The assertion coverage results are shown in Table~\ref{results1}.
We recorded the number of tests for which the requirement was satisfied (P), not satisfied (F) or not checked (NC).

\begin{table}[!t]
\centering
\renewcommand{\arraystretch}{1.3}
\caption{Requirement (assertion) coverage}
\begin{tabular}{|c|c|c|c|c|c|c|c|c|c|}
\hline 
Req. & \multicolumn{3}{|c|}{Pseudorandom} &\multicolumn{3}{|c|}{Model checking TA}  & \multicolumn{3}{|c|}{BDI agents} \\
	&  		P & F & NC		 & P & F & NC 		 & P & F & NC 		  \\
\hline
\multicolumn{10}{|c|}{Cooperative Manufacturing Assistant (160 tests per method)} \\ \hline
1 		& 19  & 4 & 137 		&  37 & 50 & 73 		& 45 & 16 & 99\\
2 		& 3  & 0 & 157		&  35 & 10 & 115		& 31 & 5 & 124\\
3 		& 24  & 0 & 136 		&  97 & 0 & 63		& 63 & 2 & 95\\
4 		& 57 & 0 & 103 		&  99 & 0 & 61 		& 72 & 0 & 88\\ 
5		& 33 & 31 & 96 		&  128 & 2 & 30  	& 79 & 4 & 77\\ 
6 		& 160 & 0 & 0 		&  160 & 0 & 0 		& 160 & 0 & 0\\ 
\hline
\multicolumn{10}{|c|}{Home Care Assistant (50 tests per method)} \\ \hline
1 		& 16 & 7 & 27 		&  19 & 11 & 20 		& 17 & 6 & 27\\
2 		& 19 & 7 & 24		&  13 & 16 & 21		& 21 & 7 & 22\\
3 		& 38 & 11 & 1 		&  29 & 19 & 2		& 40 & 8 & 2\\
4 		& 13 & 36 & 1 		&  2 & 46 & 2 		& 10 & 38 & 2\\ 
\hline
\end{tabular} \label{results1}
\end{table}

In the manufacturing scenario, Reqs.~1 to 3 are violated as the robot occasionally fails to decide whether to release a leg or not within the given time threshold. 
These failures were found mostly with the model-based tests, as expected, and in the case of Req.~3, only through tests generated based on BDI agents. 
Req.~5 is also violated occasionally, as the person's hand is allowed to be near the robot gripper when it closes. 
To improve this issue, the code could be augmented to stop the robot gripper when the hand is close and a handover is then not happening. 
Reqs.~4 and~6 are satisfied in all tests.
Seeing that violations of Req.~3 were only found through tests from BDI agents, these tests outperformed the ones generated pseudorandomly and by model checking TA, i.e.\ they were the most effective at finding requirement violations.

For the home care scenario, requirement violations were found with all test generation methods. 
If the robot collides with the dog, the collision causes the robot to fall over without recovery, which will prevent the robot from completing the current request and any subsequent ones.
This is reflected in the results of Reqs.~1, 2 and~4. 
Req.~3 is not satisfied as a velocity limit is not enforced in the motion control of the robot's base. 
As a consequence of failure in mission completion, depending on collisions with the dog, the overall assertion coverage results are low and quite similar for all the test generation methods.

\subsection{Cross-Product Functional Coverage Results} 

We expected that model-based methods would reach more cross-product items than pseudorandom test generation, i.e.\ that they would be more effective at cross-product coverage, especially for the manufacturing scenario, as 4 successful leg handovers are hard to achieve. 
Table~\ref{fig:cross-product} shows the coverage results for reachable combinations of $Human \times Robot$ behaviours as described in Section~\ref{ssc:simulator}.

The results for the manufacturing scenario show that it is difficult to reach some of the coverage points with tests from pseudorandom generation, as expected, due to the complexity of the interaction protocol to activate the robot.
Tests generated with BDI agents covered all the items, and similarly the tests generated by model checking TA, demonstrating their cross-product coverage effectiveness.

In the home care scenario, the coverage results were similar for all the three methods due to two factors. 
Firstly, the system malfunctions when collisions occur and fails to complete its mission, as explained before, leading to a low coverage of cross-product items with multiple valid requests for TIAGo. 
Secondly, both the models and the list of available requests for pseudorandom test generation constrain the amount of invalid requests for the robot to sporadic occurrences, thus increasing the generation of tests that contained valid requests.

\begin{table*}[!t]
\centering
\scriptsize
\renewcommand{\arraystretch}{1.3}
\caption{Reachable cross-product coverage}
\begin{tabular}{|c|c|c|c|c|c|}
\hline 
&$Human$ & $Robot$ 								& Pseudorandom & Model checking TA & BDI agents \\
\hline
\multicolumn{6}{|c|}{Cooperative Manufacturing Assistant (160 tests per method)} \\ \hline
1& 4 legs& $GPL=(1,1,1)$ for at least 1 leg 						& 0			&  24		& 24 \\
2& 4 legs& $GPL\neq (1,1,1)$ for at least 1 leg 					& 0 			&  30	 	& 31 \\
3& 4 legs and bored & Timed out at least once					& 2 			&  43 		& 32 \\ \hline
4& 3 legs & $GPL=(1,1,1)$ for at least 1 leg						& 0 			&  20	 	& 12 \\
5& 3 legs & $GPL \neq (1,1,1)$ for at least 1 leg				& 7 			&  38 		& 22  \\
6& 3 legs and bored & Timed out at least once					& 10 		&  38 		& 27 \\ \hline
7& 2 legs & $GPL=(1,1,1)$ for at least 1 leg 					& 2			&  13	 	& 5 \\ 
8& 2 legs & $GPL \neq (1,1,1)$ for at least 1 leg				& 6 			&  28 		& 10 \\
9& 2 legs and bored & Timed out at least once					& 9			&  11		& 7 \\ \hline
10& 1 leg & $GPL=(1,1,1)$ 										& 14 		&  11	 	& 9 \\ 
11& 1 leg & $GPL \neq (1,1,1)$ 									& 10 		&  14	 	& 14 \\ 
12& 1 leg & Timed out 											& 10 		&  2 		& 1 \\ \hline
13& 1 to 4 legs & Always timed out 								& 72			&  38		& 75 \\
14& No leg or bored & Always timed out		 					& 62 		&  2 		& 3 \\ \hline
\multicolumn{6}{|c|}{Home Care Assistant (50 tests per method)} \\ \hline
1& At least 1 $feed$ & At least 1 $feed$ 						& 5			& 9			& 6\\
2& At least 1 $clean$ & At least 1 $clean$ 						& 14			& 3			& 14\\
3& At least 1 $fridge$ & At least 1 $fridge$						& 4			& 12			& 4\\ 
4& At least 1 $sink$ & At least 1 $sink$							& 8			& 2 			& 9\\ \hline
5& At least 2 $feed$ or $clean$ & At least 2 $feed$ or $clean$ 	& 5			& 22 		& 5\\ \hline
6& Other commands & Idle											& 13			& 2			& 12\\ \hline
\end{tabular} \label{fig:cross-product}
\end{table*}

\subsection{Discussion}

\subsubsection{Exploration}

Answering our first research question, the presented results demonstrate that BDI-based tests perform as well as the ones from traditional test generation by model checking automata, and outperform the tests from pseudorandom generation, in terms of reaching high levels of code, assertion and cross-product coverage quickly, i.e.\ coverage effectiveness and efficiency. 
Also, BDI-based tests discovered requirement violations in the manufacturing task that tests from the other methods did not find, i.e.\ they were more effective at identifying failures.

\subsubsection{Performance}\label{sss:perf}

A comparison of effort to craft the different models, the resulting models' size, and the model exploration time to produce tests, for a roboticist with similar training using Jason and UPPAAL, is shown in Table~\ref{performance}.
The construction of automata in UPPAAL required a longer effort than constructing the BDI agents in Jason.
The syntax of BDI agents offers a more rational and intuitive structure, allowing the construction of an HRI protocol with less effort than specifying automata variables, guards and transitions. 
Specifying BDI belief sets is also more intuitive than specifying TCTL properties, as in the latter all the variables and states in the model need to be considered.
We limited the running time of the BDI model manually, and the time could have been further reduced. 
However, the model checking time varies depending on the properties; it is unpredictable and cannot be controlled as part of the test generation process. Note that in some cases model checking took significantly longer than exploring the BDI agents. 

\begin{table}[!t]
\centering
\renewcommand{\arraystretch}{1.3}
\caption{Performance of the model-based test generation methods}
\begin{tabular}{|c|c|c|}
\hline 
 & Model checking TA & BDI agents  \\
\hline
\multicolumn{3}{|c|}{Cooperative Manufacturing Assistant} \\ \hline
Model's lines of code			&725			&348 \\
No. states (transitions) or plans & 53 (72)	& 79\\
Modelling time					& $\approx 10.5$\ hrs  & $\approx 6$\ hrs \\
Model explor. time (min/test)	& 0.001\ s  & 5\ s \\
Model explor. time (max/test)	& 33.36\ s  & 5\ s \\ \hline
\multicolumn{3}{|c|}{Home Care Assistant} \\ \hline
Model's lines of code			&722			&131 \\
No. states (transitions) or plans & 42 (67) 	& 35\\
Modelling time					& $\approx 5.5$\ hrs  & $\approx 3$\ hrs \\
Model explor. time (min/test)	& 0.001\ s  & 1\ s \\
Model explor. time (max/test)	& 2.775\ s  & 1\ s \\ \hline
\end{tabular} \label{performance}
\end{table}

Although model checking is fully automatic, formalizing properties for test generation requires manual input and is often error prone. More research would be needed to automate the generation of properties to achieve high model coverage without manual effort. 
A higher level of automation in test generation with BDI agents can be achieved by using machine learning techniques for the selection of the best belief sets in terms of achieved coverage~\cite{MORSE2016}, at the cost of increased computational effort. 
In addition, BDI models can also be explored via model checking~\cite{BordiniMCAS}, instead of using verification agents as we propose here. This could complement our approach to achieve coverage closure.

\subsubsection{Practicality and Transferability}\label{sss:pas} 

Our results demonstrate that BDI agents are applicable to different HRI scenarios, as exemplified by our two case studies. 
BDI agents model an HRI task with human-like actions and rational reasoning. They are natural to program, by specifying plans of actions.
Compared to model checking, we do not need to formulate temporal logic reachability properties, which requires a good understanding of formal logics, and a greater amount of manual effort. 
In addition, constructing automata, such as TA, for larger case studies requires several cycles of abstraction to manage the state-space explosion problem~\cite{webster14formalshort}.

\subsubsection{Limitations}

In this paper, the two case studies serve to illustrate our comparison of using BDI agents, instead of model-checking TA, for model-based test generation. 
Industrial-sized code, and richer HRI case studies are required to further validate our results. 
Other coverage metrics could be employed to add further comparison dimensions in terms of system exploration during testing, such as FSM states, or transitions, making use of the FSM structure of some of the code.
Nonetheless, our approach is not prescriptive on structuring the code as FSMs, or on using SMACH. 
Finally, all the approaches presented in this paper implement offline test generation, i.e.\ the tests are computed before the simulation. This is suitable when the models of the system and the environment do not change. For robots that learn and adapt in changing environments, online techniques for test generation will be required.

\section{CONCLUSIONS AND FUTURE WORK} \label{sc:conclusion}

In this paper, we compared two model-based test generation approaches in the context of HRI scenarios: BDI agents and model checking automata, in terms of exploration (coverage), performance, practicality and transferability. 
We also compared both methods to pseudorandom test generation as a baseline. 
The test generation methods were applied to two case studies, a cooperative manufacturing task, and a home care scenario, for which high-level robot control code was tested in ROS and Gazebo simulators using a coverage-driven automated testbench~\cite{CDV2015,TAROS2016}.

We have found that BDI agents allow  realistic, human-like stimulus, whilst facilitating the generation of complex interactions between the robot and its environment.
Tests generated with BDI agents perform similarly to the ones generated by model checking TA in terms of reaching high coverage (code, assertions, and cross-product), and are better than the ones generated pseudorandomly. 
Also, BDI agents are easier to specify, computationally cheaper to execute, and more predictable in terms of performance when compared to model checking TA. 
In conclusion, our results clearly highlight the advantages of using BDI agents for test generation in complex HRI scenarios that require the robot code under test to be stimulated with a broad variety of realistic interaction sequences.

In the future, we plan to investigate how BDI agents can be used to interactively stimulate the robot code during simulation, generating new stimulus on the fly in direct response to a robot's observable behaviour within the test environment. We then intend to apply this online, BDI-based test generation to stress test complex systems with agency and change.

\bibliographystyle{plain}

\begin{thebibliography}{10}

\bibitem{Abdellatif2012}
Tesnim Abdellatif, Saddek Bensalem, Jacques Combaz, Lavindra {de Silva}, and
  Felix Ingrand.
\newblock Rigorous design of robot software: A formal component-based approach.
\newblock {\em Robotics and Autonomous Systems}, 60:1563--1578, 2012.

\bibitem{Alexander2015}
Rob Alexander, Heather Hawkins, and Drew Rae.
\newblock {Situation Coverage -- A Coverage Criterion for Testing Autonomous
  Robots}.
\newblock Technical report, Department of Computer Science, University of York,
  2015.

\bibitem{MORSE2016}
D.~Araiza-Illan, A.G. Pipe, and K.~Eder.
\newblock Intelligent agent-based stimulation for testing robotic software in
  human-robot interactions.
\newblock In {\em Proceedings of the 3rd Workshop on Model-Driven Robot
  Software Engineering (MORSE)}, 2016.

\bibitem{CDV2015}
D.~{Araiza-Illan}, D.~Western, K.~Eder, and A.~Pipe.
\newblock Coverage-driven verification --- an approach to verify code for
  robots that directly interact with humans.
\newblock In {\em Proc. HVC}, pages 1--16, 2015.

\bibitem{TAROS2016}
D.~{Araiza-Illan}, D.~Western, K.~Eder, and A.~Pipe.
\newblock Systematic and realistic testing in simulation of control code for
  robots in collaborative human-robot interactions.
\newblock In {\em Proc. TAROS}, 2016.

\bibitem{BordiniMCAS}
R.~H. Bordini, M.~Fisher, C.~Pardavila, and M.~Wooldridge.
\newblock Model checking {AgentSpeak}.
\newblock In {\em Proc. AAMAS}, 2003.

\bibitem{BordiniVMAPMC}
Rafael~H. Bordini, Michael Fisher, Willem Visser, and Michael Wooldridge.
\newblock Verifying multi-agent programs by model checking.
\newblock {\em Journal of Autonomous Agents and Multi-Agent Systems},
  12(2):239--256, 2006.

\bibitem{Agentspeakbook}
R.H. Bordini, J.F. H\"{u}bner, and M.~Wooldridge.
\newblock {\em Programming Multi-Agent Systems in {AgentSpeak} using {Jason}}.
\newblock Wiley, 2007.

\bibitem{SMACH}
Jonathan Boren and Steve Cousins.
\newblock {The SMACH High-Level Executive}.
\newblock {\em IEEE Robotics \& Automation Magazine}, 17(4):18--20, 2010.

\bibitem{Clarke2001}
Edmund Clarke, Orna Grumberg, Somesh Jha, Yuan Lu, , and Helmut Veith.
\newblock Progress on the state explosion problem in model checking.
\newblock In {\em Informatics. 10 Years Back. 10 Years Ahead}, number 2000 in
  LNCS, pages 176--194, 2001.

\bibitem{Clarke2003}
Edmund Clarke, Orna Grumberg, Somesh Jha, Yuan Lu, and Helmut Veith.
\newblock Counterexample-guided abstraction refinement for symbolic model
  checking.
\newblock {\em Journal of the ACM}, 50(5):752--794, 2003.

\bibitem{Dennis2016}
Louise~A. Dennis, Michael Fisher, Nicholas~K. Lincoln, Alexei Lisitsa, and
  Sandor~M. Veres.
\newblock Practical verification of decision-making in agent-based autonomous
  systems.
\newblock {\em Automated}, 23(3):305--359, 2016.

\bibitem{Diasneto2007}
Arilo~C. {Dias Neto}, Rajesh Subramanyan, Marlon Vieira, and Guilherme~H.
  Travassos.
\newblock A survey on model-based testing approaches: A systematic review.
\newblock In {\em Proc. WEASELTech}, pages 31--36, 2007.

\bibitem{ROMAN14}
K.I. Eder, C.~Harper, and U.B. Leonards.
\newblock Towards the safety of human-in-the-loop robotics: Challenges and
  opportunities for safety assurance of robotic co-workers.
\newblock In {\em Proc. IEEE ROMAN}, pages 660--665, 2014.

\bibitem{Fraser2009}
Gordon Fraser, Franz Wotawa, and Paul~E. Ammann.
\newblock Testing with model checkers: a survey.
\newblock {\em Software Testing, Verification and Reliability}, 19:215--261,
  2009.

\bibitem{Kim2006}
J.~Kim, J.~M. Esposito, and R.V. Kumar.
\newblock Sampling-based algorithm for testing and validating robot
  controllers.
\newblock {\em International Journal of Robotics Research}, 25(12):1257--1272,
  2006.

\bibitem{lenz2010bert2}
A.~Lenz, S.~Skachek, K.~Hamann, J.~Steinwender, A.G. Pipe, and C.~Melhuish.
\newblock The {BERT}2 infrastructure: {An} integrated system for the study of
  human-robot interaction.
\newblock In {\em Proc. {IEEE}-{RAS} {Humanoids}}, pages 346--351, 2010.

\bibitem{Lill2012}
R.~Lill and F.~Saglietti.
\newblock Model-based testing of autonomous systems based on {Coloured} {Petri}
  {Nets}.
\newblock In {\em Proc. ARCS}, 2012.

\bibitem{Micskei2012}
Z.~Micskei, Z.~Szatm\'{a}ri, J.~Ol\'{a}h, and I.~Majzik.
\newblock A concept for testing robustness and safety of the context-aware
  behaviour of autonomous systems.
\newblock In {\em Proc. KES-AMSTA}, pages 504–--513, 2012.

\bibitem{Mossige2014}
M.~Mossige, A.~Gotlieb, and H.~Meling.
\newblock Testing robot controllers using constraint programming and continuous
  integration.
\newblock {\em Information and Software Technology}, 57:169--185, 2014.

\bibitem{Namin2010}
Akbar~Siami Namin, Barbara Millet, and Mohan Sridharan.
\newblock Fast abstract: Stochastic model- based testing for human-robot
  interaction.
\newblock In {\em Proc. ISSRE}, 2010.

\bibitem{Nie2011}
C.~Nie and H.~Leung.
\newblock A survey of combinatorial testing.
\newblock {\em ACM Computing Surveys}, 43(2):1--29, 2011.

\bibitem{Patelli2016}
Andrea Patelli and Luca Mottola.
\newblock Model-based real-time testing of drone autopilots.
\newblock In {\em Proc. DroNet}, 2016.

\bibitem{Petters2008}
S.~Petters, D.~Thomas, M.~Friedmann, and O.~{von Stryk}.
\newblock Multilevel testing of control software for teams of autonomous mobile
  robots.
\newblock In {\em Proc. SIMPAR}, 2008.

\bibitem{Pinho2014}
T.~Pinho, A.~P. Moreira, and J.~{Boaventura-Cunha}.
\newblock Framework using {ROS} and {SimTwo} simulator for realistic test of
  mobile robot controllers.
\newblock In {\em Proc. CONTROLO}, pages 751--759, 2014.

\bibitem{Pizialli2004}
Andrew Pizialli.
\newblock {\em Functional Verification Coverage Measurement and Analysis}.
\newblock Springer, 2008.

\bibitem{Shafique2015}
Muhammad Shafique and Yvan Labiche.
\newblock A systematic review of state-based test tools.
\newblock {\em International Journal on Software Tools for Technology
  Transfer}, 17:59--76, 2015.

\bibitem{Toom2010}
A.~Toom, N.~Izerrouken, T.~Naks, M.~Pantel, and O.~Ssi {Yan Kai}.
\newblock Towards reliable code generation with an open tool: Evolutions of the
  {Gene-Auto} toolset.
\newblock In {\em Proc. ERTS}, 2010.

\bibitem{Utting2012}
M.~Utting, A.~Pretschner, and B.~Legeard.
\newblock A taxonomy of model-based testing approaches.
\newblock {\em Software Testing, Verification and Reliability}, 22:297--312,
  2012.

\bibitem{webster14formalshort}
Matt Webster, Clare Dixon, Michael Fisher, Maha Salem, Joe Saunders, Kheng~Lee
  Koay, and Kerstin Dautenhahn.
\newblock Formal verification of an autonomous personal robotic assistant.
\newblock In {\em Proc. AAAI FVHMS 2014}, pages 74--79, 2014.

\end{thebibliography}

\end{document}